\documentclass[runningheads]{llncs}
\pdfoutput=1
\usepackage{amsmath}
\usepackage{amssymb}

\titlerunning
\authorrunning

\def\ar{\leftarrow}
\def\rar{\rightarrow}

\def\beq{\begin{equation}}
\def\eeq#1{\label{#1}\end{equation}}
\def\ba{\begin{array}}
\def\ea{\end{array}}
\def\i#1{\hbox{\it #1\/}}
\def\is#1{{\hbox{\scriptsize {\it #1\/}}}}

\def\no{\i{not}}

\title{\bf On the Semantics of Gringo}
\author{Amelia Harrison, Vladimir Lifschitz, and Fangkai Yang}
\institute{University of Texas, Austin, Texas, USA \\ \{ameliaj, 
vl, fkyang\}@cs.utexas.edu} 

\titlerunning{On the Semantics of Gringo}
\authorrunning{A.~Harrison \emph{et al}\/.}
\begin{document}
\date{}
\maketitle

\begin{abstract}
Input languages of answer set solvers are based on the mathematically simple
concept of a stable model.  But many useful constructs available in these
languages, including local variables, conditional literals, and aggregates,
cannot be easily explained in terms of stable models in the sense of the
original definition of this concept and its straightforward generalizations.
Manuals written by designers of answer set solvers usually explain such
constructs using examples and informal comments that appeal to the user's
intuition, without references to any precise semantics.
We propose to approach the problem of defining the semantics of {\sc gringo}
programs by translating them into the language of infinitary propositional
formulas.  This semantics allows us to study equivalent transformations of
{\sc gringo} programs using natural deduction in infinitary propositional
logic.
\end{abstract}

\section{Introduction}

In this note, Gringo is the name of the input language of the
grounder {\sc gringo},\footnote{\tt http://potassco.sourceforge.net/.
\label{ft1}} which
is used as the front end in many answer set programming (ASP) systems.  Several
releases of {\sc gringo} have been made public, and more may be coming in
the future; accordingly, we can distinguish between several ``dialects'' of
the language Gringo.  We concentrate here on Version~4, released in March of
2013.  (It differs from Version~3, described in the {\sl User's Guide} dated
October 4, 2010,\footnote{The {\sl User's Guide} can be downloaded from the
Potassco website (Footnote~\ref{ft1}).  It is posted also at {\tt
http://www.cs.utexas.edu/users/vl/teaching/lbai/clingo\_ guide.pdf}\label{ft3}.}
in several ways, including the approach to aggregates---it is modified as
proposed by the ASP Standardization Working Group.\footnote{\tt
https://www.mat.unical.it/aspcomp2013/ASPStandardization. \label{ft2}})

The basis of Gringo is the language
of logic programs with negation as failure, with the syntax and semantics
defined in \cite{gel88}.  Our goal here is to extend that semantics to a larger
subset of Gringo.  Specifically, we would like to cover arithmetical functions
and comparisons, conditions, and aggregates.\footnote{The subset of Gringo
discussed in this note includes also constraints, disjunctive rules, and
choice rules, treated along the lines of \cite{gel91b} and \cite{fer05b}.  The
first of these papers introduces also ``classical'' (or ``strong'')
negation---a useful feature that we do not include.  (Extending our
semantics of Gringo to programs with classical negation is straightforward,
using the process of eliminating classical negation in favor of additional
atoms described in \cite[Section~4]{gel91b}.)}

Our proposal is based on the informal and sometimes incomplete
description of the language in the {\sl User's Guide}, on the discussion of
ASP programming constructs in \cite{geb12}, on experiments with
{\sc gringo}, and on the clarifications provided in response to our questions
by its designers.

The proposed semantics uses a translation from Gringo into the language
of infinitary propositional formulas---propositional formulas with infinitely
long conjunctions and disjunctions.  Including infinitary formulas is
essential, as we will see, when conditions or aggregates use variables ranging
over infinite sets (for instance, over integers). 
 
Alternatively, the semantics of Gringo can be approached using quantified 
equilibrium logic \cite{pea04} 
or its syntactic counterpart defined in \cite{fer09}. This method involves 
translating rules into the language of first-order logic. For instance, the rule
\beq
p(Y) \ar \i{count}\{X,Y:q(X,Y)\}\geq 1
\eeq{ta}
can be represented by the sentence
$$
\forall y (\exists x Q(x, y) \rar P(y)).
$$
However, this translation is not sufficiently general. For instance, it is not 
clear how to represent the rule 
\beq
\i{total\_hours}(N) \ar \i{sum}\{H,C:\i{enroll}(C),hours(H,C)\}=N
\eeq{ag-rule}
from Section 3.1.10 of the Gringo 3 {\sl User's Guide} with a first-order formula.  One 
reason is that the aggregate \i{sum} is used here instead of \i{count}. The second 
difficulty is that the variable $N$ is used rather than a constant.

General aggregate expressions, as used in Gringo,  
can be represented by first-order formulas
with generalized quantifiers.\footnote{Stable models of formulas with
generalized quantifiers are defined by Lee and Meng
\cite{lee12a}\cite{lee12b}\cite{lee12c}.} The advantage of infinitary propositional
formulas as the target language is that properties of these formulas, and of
their stable models, are better understood.  We may be able to prove, for
instance, that two Gringo programs have the same stable models by observing
that the corresponding infinitary formulas are equivalent in one of the
natural deduction systems discussed in \cite{har13}.  We give here several
examples of reasoning about Gringo programs based on this idea.

The process of converting Gringo programs into infinitary propositional
formulas defined in this note uses
substitutions to eliminate variables.  This form of grounding is
quite different, of course, from the process
of intelligent instantiation implemented in {\sc gringo} and other grounders.
Mathematically, it is much simpler than intelligent instantiation; as a
computational procedure, it is much less efficient,
not to mention the fact that sometimes it produces infinite objects.  Like
grounding in the original definition of a stable model
\cite{gel88}, it is modular, in the sense that it applies to the program
rule by rule, and it is applicable even if the program is not safe.
From this perspective, {\sc gringo}'s safety requirement is
an implementation restriction.

Our description of the syntax of Gringo disregards some of the
features related to representing programs as strings of ASCII characters,
such as using \verb+:-+ to separate the head from the body, using semicolons,
rather than parentheses, to indicate the boundaries of a conditional literal,
and representing
falsity (which we denote here by $\bot$) as \verb+#false+.  Since the subset
of Gringo discussed in this note does not include assignments, we can
disregard also the requirement that equality be represented by two characters
\verb+==+.

\section{Syntax}

We begin with a signature~$\sigma$ in the sense of first-order logic that
includes, among others,
\begin{enumerate}
\item[(i)]
numerals---object constants representing all integers,
\item[(ii)]
arithmetical functions---binary function constants $+$, $-$, $\times$,
\item[(iii)]
comparisons---binary predicate constants $<$, $>$, $\leq$, $\geq$.
\end{enumerate}
We will identify numerals with the corresponding elements of the set {\bf Z}
of integers.  Object, function,
and predicate symbols not listed under (i)--(iii) will be called
{\sl symbolic}.  A term over $\sigma$ is {\sl arithmetical} if it does not contain
symbolic object or function constants. A ground term is {\sl precomputed}
if it does not contain arithmetical functions.

We assume that in addition to the signature, a set of symbols called
{\sl aggregate names} is specified, and that for
each aggregate name~$\alpha$, {\sl the function denoted by $\alpha$}, $\widehat \alpha$,
maps every tuple of precomputed terms to an element of ${\bf Z}\cup\{\infty,-\infty\}$.

\medskip\noindent\textbf{Examples.}
The functions denoted by the aggregate names \i{count}, \i{max}, and \i{sum}
are defined as follows.  For any set~$T$ of tuples of precomputed terms,
\begin{itemize}
\item
$\widehat{\i{count}}(T)$ is the cardinality of~$T$ if~$T$ is finite, and
$\infty$ otherwise;
\item
$\widehat{\i{max}}(T)$ is the least upper bound of the set of the
integers~$t_1$ over all tuples $(t_1,\dots,t_m)$ from $T$ in which~$t_1$
is an integer;
\item
$\widehat{\i{sum}}(T)$ is the sum of the integers~$t_1$ over all
tuples $(t_1,\dots,t_m)$ from  $T$ in which~$t_1$ is a positive integer;
it is $\infty$ if there are infinitely many such tuples.\footnote{To
allow negative numbers in this example, we would have to define summation
for a set that contains both infinitely many positive numbers
and infinitely many negative numbers.  It is unclear how to do this in a 
natural way.}
\end{itemize}

\medskip
A {\sl literal} is an expression of one of the forms
$$
p(t_1,\dots,t_k),\ t_1=t_2,\ \no\ p(t_1,\dots,t_k),\ \no\ (t_1=t_2)
$$
where~$p$ is a symbolic predicate constant of arity~$k$, and each~$t_i$
is a term over~$\sigma$, or
$$t_1\prec t_2,\ \no\ (t_1\prec t_2)$$
where~$\prec$ is a comparison, and $t_1$, $t_2$ are arithmetical terms.
A {\sl conditional literal} is an expression of the form $H : {\bf L}$,
where $H$ is a literal or the symbol $\bot$, and
{\bf L} is a list of literals, possibly empty.  The members of {\bf L}
will be called {\sl conditions}.  If {\bf L} is empty then we will drop
the colon after~$H$, so that every literal can be viewed as a
conditional literal.

\medskip\noindent\textbf{Example.}
If \i{available} and \i{person} are unary predicate symbols then
$$\i{available}(X) : \i{person}(X)$$
and
$$\bot : (\i{person}(X), \no\ \i{available}(X))$$
are conditional literals.
\medskip

An {\sl aggregate expression} is an expression of the form
$$
\alpha\{{\bf t} : {\bf L} \}\prec s
$$
where $\alpha$ is an aggregate name,
${\bf t}$ is a list of terms,
${\bf L}$ is a list of literals,
$\prec$ is a comparison or the symbol~$=$,
and $s$ is an arithmetical term.

\medskip\noindent\textbf{Example.}
If \i{enroll} is a unary predicate symbol and \i{hours} is a binary
predicate symbol then
$$
\i{sum}\{H,C:\i{enroll}(C),\i{hours}(H,C)\}=N
$$
is an aggregate expression.

\medskip
A {\sl rule} is an expression of the form
\beq
H_1\,|\,\cdots\,|\,H_m \ar B_1,\dots, B_n
\eeq{rule3}
($m, n \geq 0$), where each~$H_i$ is a conditional literal, and each~$B_i$
is a conditional literal or an aggregate expression.
A {\sl program} is a set of rules.

If~$p$ is a symbolic predicate constant of arity~$k$, and~{\bf t} is a
$k$-tuple of terms, then
$$\{p({\bf t})\} \ar B_1,\dots, B_n$$
is shorthand for
$$p({\bf t})\ |\ \no\ p({\bf t}) \ar B_1,\dots, B_n.$$

\medskip\noindent\textbf{Example.}
For any positive integer~$n$,
\beq
\ba {rl}
\{p(i)\}\!\!&\ar\qquad\qquad\qquad\qquad\qquad\qquad(i=1,\dots,n),\\
\!\!&\ar p(X), p(Y), p(X\!+\!Y)
\ea
\eeq{ex1}
is a program.

\section{Semantics}

We will define the semantics of Gringo using a syntactic transformation~$\tau$.
It converts Gringo rules into infinitary propositional combinations of
atoms of the form $p({\bf t})$, where~$p$ is a symbolic predicate constant,
and~{\bf t} is a tuple of precomputed terms.  
Then the stable models of a program will be defined as stable models, in the 
sense of \cite{tru12},   
of the set consisting of the translations of all rules of the program.
Truszczynski's definition of stable models for infinitary propositional 
formulas is reviewed below.

Prior to defining the translation $\tau$ for rules, we will define it 
for ground literals, conditional literals, and aggregate expressions. 

\subsection{Review: Stable Models of Infinitary Formulas} \label{sec:sm}

Let $\sigma$ be a propositional signature,
that is, a set of propositional atoms.  The sets
$\mathcal{F}^\sigma_0$,
$\mathcal{F}^\sigma_1$, $\ldots$ are defined as follows:
\begin{itemize}
\item $\mathcal{F}^\sigma_0=\sigma\cup\{\bot\}$,
\item $\mathcal{F}^\sigma_{i+1}$ is obtained from $\mathcal{F}^\sigma_{i}$ by 
adding expressions $\mathcal{H}^\land$ and $\mathcal{H}^\lor$ for all subsets 
$\mathcal{H}$ of $\mathcal{F}^\sigma_i$, and expressions $F\rar G$ for all 
$F,G\in\mathcal{F}^\sigma_i$.
\end{itemize}
The elements of $\bigcup^{\infty}_{i=0}\mathcal{F}^\sigma_i$ are called {\it 
(infinitary) formulas} over $\sigma$. 
Negation and equivalence are abbreviations. 

Subsets of a signature~$\sigma$ will be also called its {\it interpretations}.
The satisfaction relation between an interpretation and a formula is 
defined in a natural way. 

The {\it reduct} $F^I$ of a formula~$F$ w.r.t. an
interpretation~$I$ is defined as follows:
\begin{itemize}
\item $\bot^I=\bot$.
\item For $p\in \sigma$, $p^I=\bot$ if $I\not\models p$; otherwise $p^I=p$.
\item $(\mathcal{H}^\land)^I=\{G^I\ |\ G\in\mathcal{H}\}^\land$.
\item $(\mathcal{H}^\lor)^I=\{G^I\ |\ G\in\mathcal{H}\}^\lor$.
\item $(G\rar H)^I=\bot$ if $I\not\models G\rar H$; otherwise $(G\rar
H)^I=G^I\rar H^I$.
\end{itemize}
An interpretation~$I$ is a {\it stable model} of a set $\mathcal{H}$ of  
formulas if it is minimal w.r.t. set inclusion among the interpretations
satisfying the reducts of all formulas from~$\mathcal{H}$.

\subsection{Semantics of Well-Formed Ground Literals}

A term~{\bf t} is {\sl well-formed} if it contains neither symbolic object
constants nor symbolic function constants in the scope of arithmetical
functions.  For instance, all arithmetical terms and all precomputed terms
are well-formed; $c\!+\!2$ is not well-formed.  The definition of
``well-formed'' for literals, aggregate expressions, and so forth is the
same.

For every well-formed ground term~$t$, by $[t]$ we denote the precomputed
term obtained from~$t$ by evaluating all arithmetical functions, and
similarly for tuples of terms.  For instance, $[f(2\!+\!2)]$ is $f(4)$.

The translation~$\tau L$ of a well-formed ground literal~$L$ is defined as
follows:
\begin{itemize}
\item
$\tau (p({\bf t}))$ is $p([{\bf t}])$;
\item
$\tau(t_1\prec t_2)$, where $\prec$ is the symbol $=$ or a comparison,
is~$\top$ if the relation~$\prec$ holds between $[t_1]$ and $[t_2]$, and~$\bot$
otherwise;
\item
$\tau(\no\ A)$ is $\neg \tau A$.
\end{itemize}
For instance, $\tau(\no\ p(f(2\!+\!2)))$ is $\neg p(f(4))$, and
$\tau (2\!+\!2\!=4)$ is~$\top$.

Furthermore, $\tau\bot$ stands for ~$\bot$, and, for any list {\bf L} of
ground literals, $\tau{\bf L}$ is the conjunction of the formulas $\tau L$
for all members~$L$ of~{\bf L}.

\subsection{Global Variables}

About a variable we say that it is {\sl global}
\begin{itemize}
\item
in a conditional literal $H : {\bf L}$, if it occurs in~$H$ but does not
occur in~{\bf L};
\item
in an aggregate expression~$\alpha\{{\bf t} : {\bf L} \}\prec s$, if it
occurs in the term~$s$;
\item
in a rule~(\ref{rule3}), if it is global in at least one of the
expressions~$H_i$,~$B_i$.
\end{itemize}
For instance, the head of the rule (\ref{ag-rule})
is a literal with the global variable~$N$, and its body is an aggregate
expression with the global variable~$N$.  Consequently~$N$ is global in the
rule as well.

A conditional literal, an aggregate expression, or a rule is {\sl closed} if
it has no global variables.  An {\sl instance} of a rule~$R$ is any well-formed
closed rule that can be obtained from~$R$ by substituting precomputed terms
for global variables.  For instance,
$$\i{total\_hours}(6) \ar \i{sum}\{H,C:\i{enroll}(C),hours(H,C)\}=6$$
is an instance of rule~(\ref{ag-rule}).  It is clear that if a rule is not
well-formed then it has no instances.

\subsection{Semantics of Closed Conditional Literals}

If~$t$ is a term,~{\bf x} is a tuple of distinct variables, and~{\bf r} is
a tuple of terms of the same length as~{\bf x}, then the term obtained
from~$t$ by substituting~{\bf r}
for~{\bf x} will be denoted by $t^{\bf x}_{\bf r}$.  Similar notation will be
used for the result of substituting~{\bf r} for~{\bf x} in expressions of
other kinds, such as literals and lists of literals.

The result of applying~$\tau$ to a closed conditional literal~$H : {\bf L}$
is the conjunction of the formulas
$$\tau ({\bf L}^{\bf x}_{\bf r})\rar \tau (H^{\bf x}_{\bf r})$$
where {\bf x} is the list of variables occurring in~$H : {\bf L}$,
over all tuples {\bf r} of precomputed terms of the same length as~{\bf x}
such that both ${\bf L}^{\bf x}_{\bf r}$ and $H^{\bf x}_{\bf r}$ are well-formed.
For instance, 
$$\tau(\i{available}(X) : \i{person}(X))$$ 
is the conjunction of the formulas $\i{person}(r)\rar\i{available}(r)$ over all 
precomputed terms~$r$; $$\tau(\bot:p(2\times X))$$ is the conjunction of the 
formulas $\neg p(2\times i)$ over all numerals~$i$.  When a conditional literal 
occurs in the head of a rule, we will translate it in a different way.  By 
$\tau_h(H : {\bf L})$ we denote the disjunction of the formulas 
$$\tau ({\bf L}^{\bf x}_{\bf r})\land \tau (H^{\bf x}_{\bf r})$$ where {\bf x} 
and {\bf r} are as above.  For instance, $$\tau_h(\i{available}(X) : 
\i{person}(X))$$ is the disjunction of the formulas $\i{person}(r)\land 
\i{available}(r)$ over all precomputed terms~$r$.  

\subsection{Semantics of Closed Aggregate Expressions}

In this section, the semantics of ground aggregates proposed in 
\cite[Section~4.1]{fer05} is adapted to closed aggregate expressions.  Let~$E$ 
be a closed aggregate expression~$\alpha\{{\bf t} : {\bf L} \}\prec s$, and 
let~{\bf x} be the list of variables occurring in~$E$.  A tuple~{\bf r} of 
precomputed terms of the same length as~{\bf x} is {\sl admissible} (w.r.t.~$E$) 
if both ${\bf t}^{\bf x}_{\bf r}$ and ${\bf L}^{\bf x}_{\bf r}$ are 
well-formed.  About a set~$\Delta$ of admissible tuples we say that it 
{\sl justifies}~$E$ if the relation $\prec$ holds between 
$\widehat\alpha(\{[{\bf t}^{\bf x}_{\bf r}] : {\bf r}\in\Delta\})$ and $[s]$.  
For instance, consider the aggregate expression 
\beq 
\i{sum}\{H,C:\i{enroll}(C), hours(H,C)\}=6.  
\eeq{ag-ex} 
In this case, admissible tuples are arbitrary pairs of precomputed terms.  
The set $\{(3,\i{cs101}),(3,\i{cs102})\}$ justifies~(\ref{ag-ex}), because 
$$\widehat{\i{sum}}(\{(H,C)^{H,C}_{3,\is{cs101}},(H,C)^{H,C}_{3,\is{cs102}}\}) 
=\widehat{\i{sum}}(\{(3,\i{cs101}),(3,\i{cs102})\})=3+3=6.$$ More generally, a 
set~$\Delta$ of pairs of precomputed terms
justifies~(\ref{ag-ex}) whenever~$\Delta$ contains finitely many
pairs~$(h,c)$ in which~$h$ is a positive integer, and the sum of the
integers~$h$ over all these pairs is~6.

We define $\tau E$ as the conjunction of the implications
\beq
    \bigwedge_{{\bf r}\in\Delta}\tau({\bf L}^{\bf x}_{\bf r})
\rar\bigvee_{{\bf r}\in A\setminus\Delta}\tau({\bf L}^{\bf x}_{\bf r})
\eeq{agsem}
 over all sets~$\Delta$ of
admissible tuples that do not justify~$E$, where $A$ is the set of all 
admissible tuples.
For instance, if~$E$ is~(\ref{ag-ex}) then the conjunctive terms of~$\tau E$
are the formulas
$$
    \bigwedge_{(h,c)\in\Delta}(\i{enroll}(c)\land\i{hours}(h,c))
\rar\bigvee_{(h,c)\not\in\Delta}(\i{enroll}(c)\land\i{hours}(h,c)).
$$
The conjunctive term corresponding to $\{(3,\i{cs101})\}$ as $\Delta$ says:
if I am enrolled in CS101 for 3 hours then I am enrolled in at least one
other course.

\subsection{Semantics of Rules and Programs} \label{ssec:grules}

For any rule~$R$, $\tau R$ stands for the conjunction of the formulas
$$
\tau B_1\land\cdots\land\tau B_n\rar\tau_h H_1\lor\cdots\lor\tau_h H_m
$$
for all instances~(\ref{rule3}) of~$R$.
A {\sl stable model} of a program~$\Pi$ is a stable model, in the sense
of~\cite{tru12}, of the set consisting of the formulas~$\tau R$ for all
rules~$R$ of~$\Pi$.

Consider, for instance, the rules of program~(\ref{ex1}).  If~$R$ is the
rule~$\{p(i)\}$ then $\tau R$ is
\beq
p(i)\lor\neg p(i)
\eeq{r1}
($i=1,\dots,n$). If~$R$ is the rule
$$\ar p(X), p(Y), p(X\!+\!Y)$$
then the instances of~$R$ are rules of the form
$$\ar p(i), p(j), p(i\!+\!j)$$
for all numerals $i$, $j$.  (Substituting precomputed ground terms other
than numerals would produce a rule that is not well-formed.)  Consequently
$\tau R$ is in this case the infinite conjunction
\beq
\bigwedge_{{i,j,k\in{\bf Z}}\atop{i+j=k}}\neg(p(i)\land p(j)\land p(k)).
\eeq{r2}
The stable models of program~(\ref{ex1}) are the stable models of
formulas~(\ref{r1}),~(\ref{r2}), that is, sets of the form
$\{p(i) : i\in S\}$ for all sum-free subsets~$S$ of $\{1,\dots,n\}$.

\section{Reasoning about Gringo Programs}

In this section we give examples of reasoning about Gringo programs on the
basis of the semantics defined above.  These examples use the results of
\cite{har13}, and we assume here that the reader is familiar with that paper.

\subsection{Simplifying a Rule from Example 3.7 of User's Guide}

Consider the rule\footnote{This rule is similar to a rule from Example~3.7 of 
the Gringo 3 {\sl User's Guide} (see Footnote~\ref{ft3}).}
\beq
  \i{weekdays} \ar \i{day}(X) : (\i{day}(X) , \no\ \i{weekend}(X)). 
\eeq{ex2}
Replacing this rule with the fact \i{weekdays} within any program will not
affect the set of stable models.  Indeed, the result of applying
translation~$\tau$ to~(\ref{ex2}) is the formula
\beq
\bigwedge_{r}(\i{day}(r) \land \neg\i{weekend}(r) \rar \i{day}(r)) \,\rar\,
\i{weekdays},
\eeq{ex2a}
where the conjunction extends over all precomputed terms~$r$.
The formula
$$\i{day}(r) \land \neg\i{weekend}(r) \rar \i{day}(r)$$
is intuitionistically provable.  By the replacement property of the basic
system of natural deduction from \cite{har13}, it follows that~(\ref{ex2a})
is equivalent to \i{weekdays} in the basic system.  By the main theorem
of \cite{har13}, it follows that replacing (\ref{ex2a}) with the atom
\i{weekdays} within any set of formulas does not affect the set of stable
models.

\subsection{Simplifying the Sorting Rule}

The rule
\beq
\i{order}(X,Y) \ar p(X),\, p(Y),\, X<Y,\, \no\ p(Z) : (p(Z) , X<Z , Z<Y)
\eeq{ex_sort}
can be used for sorting.\footnote{This rule was communicated to
us by Roland Kaminski on October 21, 2012.} It can be replaced by either of 
the following two shorter rules within any program without changing that 
program's stable models.  
\beq
\i{order}(X,Y) \ar p(X),\, p(Y),\, X<Y,\, \bot : (p(Z) , X<Z , Z<Y)
\eeq{eq1}
\beq
\;\;\,\i{order}(X,Y) \ar p(X),\, p(Y),\, X<Y,\, \no\ p(Z) : (X<Z , Z<Y)
\eeq{eq2}

Let's prove this claim for rule (\ref{eq1}). 
By the main theorem of \cite{har13} it is sufficient to show that the result
of applying $\tau$ to (\ref{ex_sort}) is equivalent in the basic system to
the result of applying $\tau$ to (\ref{eq1}).
The instances of (\ref{ex_sort}) are the rules
$$
\i{order}(i,j) \ar p(i),\, p(j),\, i<j,\, \no\ p(Z) : (p(Z) , i<Z , Z<j),
$$
and the instances of (\ref{eq1}) are the rules
$$
\i{order}(i,j) \ar p(i),\, p(j),\, i<j,\, \bot : (p(Z) , i<Z , Z<j)
$$
where $i$ and $j$ are arbitrary numerals.
The result of applying $\tau$ to (\ref{ex_sort}) is  the conjunction of
the formulas 
\beq
p(i) \land p(j) \land i<j \land \bigwedge_{k} \left( \neg p(k) \land i<k \land k<j \rar p(k)
\right)  \rar \i{order}(i,j) 
\eeq{}
for all numerals $i,\ j$.
The result of applying $\tau$ to (\ref{eq1}) is  the conjunction of
the formulas 
\beq
p(i) \land p(j) \land i<j \land \bigwedge_{k} \left( \neg p(k) \land i<k \land k<j \rar \bot 
\right)  \rar \i{order}(i,j). 
\eeq{}
By the replacement property of the basic system, it is sufficient to observe
 that $$p(k) \land  i<k 
\land k<j \rar \neg p(k)$$ is intuitionistically equivalent to $$p(k) \land 
i < k \land k<j \rar \bot.$$

The proof for rule (\ref{eq2}) is similar.  Rule (\ref{eq1}), like rule
(\ref{ex_sort}), is safe; rule (\ref{eq2}) is not.
 
\subsection{Eliminating Choice in Favor of a Conditional Literal}

Replacing the rule
\beq
\{p(X)\} \ar q(X)
\eeq{choice}
with
\beq
p(X) \ar q(X), \bot : \no\ p(X)
\eeq{choice_s}
within any program will not affect the set of stable models.  Indeed, the
result of applying translation~$\tau$ to~(\ref{choice}) is
\beq
\bigwedge_r(q(r)\rar p(r) \lor\neg p(r))
\eeq{c1}
where the conjunction extends over all precomputed terms~$r$, and the
result of applying~$\tau$ to~(\ref{choice_s}) is
\beq
\bigwedge_r(q(r)\land\neg\neg p(r)\rar p(r)).
\eeq{c2}
The implication from~(\ref{c1}) is equivalent to the implication from~(\ref{c2})
in the extension of intuitionistic logic obtained by adding the axiom schema
$$\neg F\lor\neg\neg F,$$
and consequently in the extended system
presented in \cite[Section 7]{har13}.   By the replacement property of the 
extended system, it follows that~(\ref{c1}) is equivalent to
(\ref{c2}) in the extended system as well. 

\subsection{Eliminating a Trivial Aggregate Expression}

The rule (\ref{ta})
says, informally speaking, that we can conclude $p(Y)$ once we established
that there exists at least one~$X$ such that $q(X,Y)$.  Replacing this rule
with
\beq
p(Y) \ar q(X,Y)
\eeq{tas}
within any program will not affect the set of stable models.

To prove this claim, we need to calculate the result of applying~$\tau$
to rule~(\ref{ta}).  The instances of~(\ref{ta}) are the rules
\beq
p(t) \ar \i{count}\{X,t:q(X,t)\}\geq 1
\eeq{tai}
for all precomputed terms~$t$.  Consider the aggregate expression~$E$
in the body of~(\ref{tai}). Any precomputed term~$r$ is admissible
w.r.t.~$E$.  A set~$\Delta$ of precomputed terms justifies~$E$ if
$$\widehat{\i{count}}(\{(r,t):r\in\Delta\})\geq 1,$$
that is to say, if~$\Delta$ is non-empty.  Consequently $\tau E$
consists of only one implication~(\ref{agsem}), with the empty~$\Delta$.
The antecedent of this implication is the empty conjunction~$\top$, and
its consequent is the disjunction $\bigvee_uq(u,t)$ over all precomputed
terms~$u$.  Then the result of applying~$\tau$ to~(\ref{ta}) is
\beq
\bigwedge_t\left(\bigvee_uq(u,t)\;\rar\; p(t)\right).
\eeq{tau_ta}
On the other hand, the result of applying~$\tau$ to~(\ref{tas}) is
$$\bigwedge_{t,u}(q(u,t)\rar p(t)).$$
This formula is equivalent to~(\ref{tau_ta}) in the basic system
\cite[Example~2]{har13}.

\subsection{Replacing an Aggregate Expression with a Conditional Literal}

Informally speaking, the rule
\beq
q \ar \i{count}\{X:p(X)\} = 0
\eeq{ar1}
says that we can conclude $q$ once we have established that the 
cardinality of the set $\{X:p(X)\}$ is $0$; the rule 
\beq
q \ar \bot:p(X)
\eeq{cl1}
says that we can conclude $q$ once we have established that $p(X)$
does not hold for any $X$. 
We'll prove that replacing (\ref{ar1}) with (\ref{cl1})
within any program will not affect the set of stable models. 
To this end, we'll show that the results of applying $\tau$
to (\ref{ar1}) and (\ref{cl1}) are equivalent to each other
in the extended system from \cite[Section 7]{har13}. 

First, we'll need to calculate the result of applying~$\tau$
to rule~(\ref{ar1}). Consider the aggregate expression~$E$ in the body
of~(\ref{ar1}). Any precomputed term~$r$ is admissible w.r.t.~$E$. 
A set~$\Delta$ of precomputed terms justifies~$E$ if
$$\widehat{\i{count}}(\{r:r\in\Delta\})=0,$$
that is to say, if~$\Delta$ is empty. Consequently $\tau E$
is the conjunction of the implications  
\beq
\bigwedge_{r \in
\Delta} p(r)  \rar  \bigvee_{r \in A \setminus \Delta} p(r)
\eeq{ar2}
for all non-empty subsets $\Delta$ of the set $A$ of precomputed terms. 
The result of applying $\tau$ to (\ref{ar1}) is 
\beq
\left ( \bigwedge_{\Delta \subseteq A \atop \Delta \not = \emptyset} 
\left ( \bigwedge_{r \in
\Delta} p(r)  \rar  \bigvee_{r \in A \setminus \Delta} p(r) \right ) \right )
\rar q.
\eeq{ar3}
The result of applying $\tau$ to (\ref{cl1}), on the other hand, is
\beq
\left ( \bigwedge_{r \in A} \neg p(r) \right ) \rar q.
\eeq{cl2}
The fact that the antecedents of (\ref{ar3}) and (\ref{cl2}) are 
equivalent to each other in the extended system can be 
established by essentially the same argument as in 
\cite[Example 7]{har13}. By the replacement property of the 
extended system, it follows that~(\ref{ar3}) is equivalent to
(\ref{cl2}) in the extended system as well. 

\subsection{Eliminating Summation over the Empty Set}

Informally speaking, the rule
\beq
q \ar \i{sum}\{X:p(X)\} = 0
\eeq{ar32}
says that we can conclude $q$ once we have established that the 
sum of the elements of the set $\{X:p(X)\}$ is $0$. In the presence of 
the constraint 
\beq
\ar p(X),
\eeq{cl3}
replacing (\ref{ar32}) with the fact $q$ will not affect the stable models. 

To see this, first we calculate the result of applying~$\tau$
to rule~(\ref{ar32}). Consider the aggregate expression~$E$ in the body
of~(\ref{ar32}). Any precomputed term~$r$ is admissible w.r.t.~$E$. 
A set~$\Delta$ of precomputed terms justifies~$E$ if
$$\widehat{\i{sum}}(\{r:r\in\Delta\})=0,$$
that is to say, if~$\Delta$ contains no positive integers. 
Consequently $\tau E$ is the conjunction of the implications  
\beq
\bigwedge_{r \in
\Delta} p(r)  \rar  \bigvee_{r \in A \setminus \Delta} p(r)
\eeq{ar4}
for subsets $\Delta$ of the set $A$ of precomputed terms that  
contain at least one positive integer. 
The result of applying $\tau$ to (\ref{ar32}) is 
\beq
\left ( \bigwedge_{\Delta \subseteq A \atop \Delta \cap {\bf Z} \not = \emptyset} 
\left ( \bigwedge_{r \in
\Delta} p(r)  \rar  \bigvee_{r \in A \setminus \Delta} p(r) \right ) \right )
\rar q.
\eeq{ar5}
The result of applying $\tau$ to (\ref{cl3}), on the other hand, is
\beq
\bigwedge_{r \in A} \neg p(r).
\eeq{cl4}
For every nonempty $\Delta$, the antecedent of (\ref{ar4}) contradicts
(\ref{cl4}). Consequently, the antecedent of (\ref{ar5}) can be derived from
(\ref{cl4}) in the basic system. It follows that the equivalence between (\ref{ar5}) and the atom $q$ can be derived in the basic system under assumption (\ref{cl4}). 

\section{Conclusion}

In this note we approached the problem of defining the semantics
of Gringo by reducing Gringo programs to infinitary
propositional formulas.  We argued that this approach to semantics
may allow us to study equivalent transformations of programs
using natural deduction in infinitary propositional logic.

In the absence of a precise semantics, it is
impossible to put the study of some important issues on a firm foundation.
This includes the correctness of ASP programs, grounders, solvers, and
optimization methods, and also the relationship between input languages of
different solvers (for instance, the equivalence of the semantics of
aggregate expressions in Gringo to their semantics in the ASP Core
language and in the language proposed in \cite{gel02} under the assumption
that aggregates are used nonrecursively).  As future work, we are 
interested in addressing some of these tasks on the basis of the semantics
proposed in this note. Proving the correctness of the intelligent instantiation
algorithms implemented in {\sc gringo} will provide justification for our 
informal claim that for a safe program, the semantics proposed here correctly 
describes the output produced by {\sc gringo}.

\section*{Acknowledgements}

Many thanks to Roland Kaminski and Torsten Schaub for helping us understand
the input language of {\sc gringo}.  Roland, Michael Gelfond, Yuliya Lierler,
Joohyung Lee, and anonymous referees provided valuable comments on drafts of 
this note.

\bibliographystyle{splncs03}
\bibliography{bib}

\begin{thebibliography}{10}
\providecommand{\url}[1]{\texttt{#1}}
\providecommand{\urlprefix}{URL }

\bibitem{fer05}
Ferraris, P.: Answer sets for propositional theories. In: Proceedings of
  International Conference on Logic Programming and Nonmonotonic Reasoning
  ({LPNMR}). pp. 119--131 (2005)

\bibitem{fer09}
Ferraris, P., Lee, J., Lifschitz, V.: Stable models and circumscription.
  Artificial Intelligence  175,  236--263 (2011)

\bibitem{fer05b}
Ferraris, P., Lifschitz, V.: Weight constraints as nested expressions. Theory
  and Practice of Logic Programming  5,  45--74 (2005)

\bibitem{geb12}
Gebser, M., Kaminski, R., Kaufmann, B., Schaub, T.: Answer Set Solving in
  Practice. Synthesis Lectures on Artificial Intelligence and Machine Learning,
  Morgan and Claypool Publishers (2012)

\bibitem{gel02}
Gelfond, M.: Representing knowledge in {A}-{P}rolog. Lecture Notes in Computer
  Science  2408,  413--451 (2002)

\bibitem{gel88}
Gelfond, M., Lifschitz, V.: The stable model semantics for logic programming.
  In: Kowalski, R., Bowen, K. (eds.) Proceedings of International Logic
  Programming Conference and Symposium. pp. 1070--1080. MIT Press (1988)

\bibitem{gel91b}
Gelfond, M., Lifschitz, V.: Classical negation in logic programs and
  disjunctive databases. New Generation Computing  9,  365--385 (1991)

\bibitem{har13}
Harrison, A., Lifschitz, V., Truszczynski, M.: On equivalent transformations of
  infinitary formulas under the stable model semantics (preliminary
  report)\footnote{\tt http://www.cs.utexas.edu/users/vl/papers/etinf.pdf}. In:
  Proceedings of International Conference on Logic Programming and Nonmonotonic
  Reasoning (LPNMR) (2013), to appear

\bibitem{lee12a}
Lee, J., Meng, Y.: Stable models of formulas with generalized quantifiers. In:
  Working Notes of the 14th International Workshop on Non-Monotonic Reasoning
  (NMR) (2012)

\bibitem{lee12b}
Lee, J., Meng, Y.: Stable models of formulas with generalized quantifiers
  (preliminary report). In: Technical Communications of the 28th International
  Conference on Logic Programming (ICLP). pp. 61--71 (2012)

\bibitem{lee12c}
Lee, J., Meng, Y.: Two new definitions of stable models of logic programs with
  generalized quantifiers. In: Working Notes of the 5th Workshop of Answer Set
  Programming and Other Computing Paradigms (ASPOCP) (2012)

\bibitem{pea04}
Pearce, D., Valverde, A.: Towards a first order equilibrium logic for
  nonmonotonic reasoning. In: Proceedings of {E}uropean Conference on Logics in
  Artificial Intelligence ({JELIA}). pp. 147--160 (2004)

\bibitem{tru12}
Truszczynski, M.: Connecting first-order {ASP} and the logic {FO(ID)} through
  reducts. In: Correct Reasoning: Essays on Logic-Based AI in Honor of Vladimir
  Lifschitz. Springer (2012)

\end{thebibliography}

\end{document}